\documentclass[letterpaper, 10 pt, conference]{ieeeconf}  
\usepackage{array} 
\usepackage{amsmath} 
\usepackage{amssymb}
\usepackage{algorithm}
\usepackage{algpseudocode}
\usepackage{booktabs} 
\usepackage{cite}
\usepackage{enumerate}
\usepackage{graphicx} 
\usepackage{multirow}
\usepackage{amsmath}
\usepackage[hidelinks]{hyperref}

\IEEEoverridecommandlockouts                              

\title{\LARGE \bf
Semantic-Geometric-Physical-Driven Robot Manipulation Skill Transfer via Skill Library and Tactile Representation
}

\author{Mingchao Qi, Yuanjin Li, Xing Liu$^{*}$, Zhengxiong Liu, Panfeng Huang
	\thanks{Mingchao Qi and Yuanjin Li contributed equally to this work.}
\thanks{This work was supported by the National Natural Science Foundation of China under Grant 92370123, 62103334, and 62273280.}
\thanks{Mingchao Qi, Yuanjin Li, Xing Liu (Corresponding author), Zhengxiong Liu, and Panfeng Huang are with the Research Center for Intelligent Robotics, School of Astronautics, Northwestern Polytechnical
	University, Xi’an, China, 710072.
        {\tt\small xingliu@nwpu.edu.cn}}%
}

\begin{document}

\maketitle
\thispagestyle{empty}
\pagestyle{empty}

\begin{abstract}
Developing general robotic systems capable of manipulating in unstructured environments is a significant challenge, particularly as the tasks involved are typically long-horizon and rich-contact, requiring efficient skill transfer across different task scenarios. To address these challenges, we propose knowledge graph-based skill library construction method. This method hierarchically organizes manipulation knowledge using ``task graph'' and ``scene graph'' to represent task-specific and scene-specific information, respectively. Additionally, we introduce ``state graph'' to facilitate the interaction between high-level task planning and low-level scene information.
Building upon this foundation, we further propose a novel hierarchical skill transfer framework based on the skill library and tactile representation, which integrates high-level reasoning for skill transfer and low-level precision for execution. At the task level, we utilize large language models (LLMs) and combine contextual learning with a four-stage chain-of-thought prompting paradigm to achieve subtask sequence transfer. At the motion level, we develop an adaptive trajectory transfer method based on the skill library and the heuristic path planning algorithm. At the physical level, we propose an adaptive contour extraction and posture perception method based on tactile representation. This method dynamically acquires high-precision contour and posture information from visual-tactile images, adjusting parameters such as contact position and posture to ensure the effectiveness of transferred skills in new environments. Experiments demonstrate the skill transfer and adaptability capabilities of the proposed methods across different task scenarios. Project website: \url{https://github.com/MingchaoQi/skill_transfer}
\end{abstract}


\section{INTRODUCTION}

Building a universal robotic manipulating system has always been a challenging task, mainly due to the complexity and variability of the real world \cite{huang2024rekep, wu2023learning, wu2024canonical}. Especially when faced with long-horizon, contact-rich complex tasks, it is difficult to program the robot for each specific scenario \cite{heo2023furniturebench}. A promising solution is to enable the skills learned by robots in previous scenarios to have good generalization, allowing efficient transfer to similar new scenarios \cite{pan2025omnimanip}. For example, tasks such as opening various types of drawers or cabinets to retrieve and place objects, or assembling different parts in a factory in distinct ways. In these scenarios, a similar set of robot skill sequences is required. However, it is important to note that the objects involved and the motion types may differ.
\begin{figure}[htbp]
    \centering
    \includegraphics[width=\linewidth]{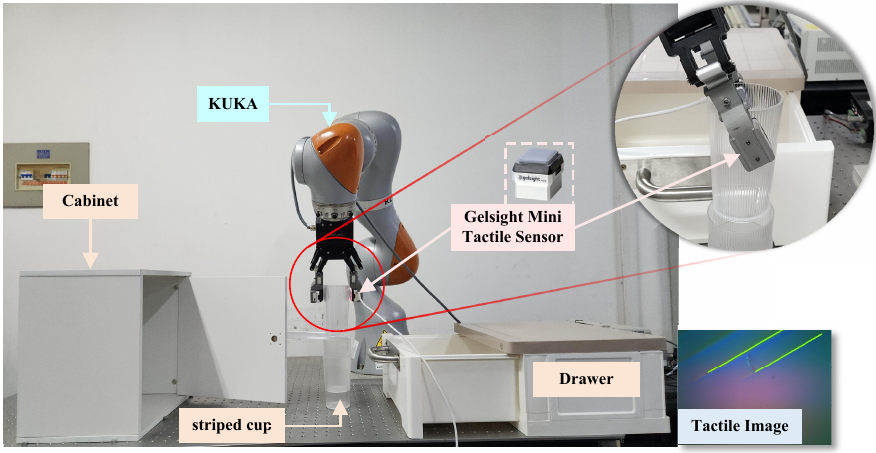} 
    \caption{Skill Transfer for Manipulation. Transferring skills learned from previous scenarios to new contexts is essential. For example, the skill of opening a drawer to retrieve and stack cups should be transferred to the new task of opening a door to retrieve cups in different orientations and complete the stacking.}
    \label{experiment_setup}
\end{figure}

One way to enhance manipulation generalization is through imitation learning, where large-scale robot datasets are used to pre-train Vision-Language Models (VLM) to construct Vision-Language Action Models (VLA) \cite{brohan2022rt}, \cite{duan2024manipulate}. These can learn strong multimodal policies, but obtaining diverse, high-quality robotic data is costly and time-consuming, and the method’s generalization capability remains limited when handling new scenes, especially with unseen object states. An alternative way to enhance manipulation generalization is to apply the world knowledge embedded in Large Language Models (LLM) to policy learning \cite{tang2023graspgpt}, \cite{huang2024copa}, reducing data dependency. Some research utilizes LLM’s reasoning ability to predict reasonable action sequences and parse environmental information into text representations \cite{mirchandani2023large}, \cite{mandi2024roco}. However, since these methods rely on coarse-grained environmental representations, they only manage simple tasks like pick-and-place and object rearrangement. Particularly for long-horizon tasks, LLMs struggle to rationally plan fully executable action sequences. Moreover, these methods often focus on the combination of high-level skills while neglecting the properties of manipulation objects and underlying contact issues, limiting skill generalization in contact-rich tasks.

To address these issues, we propose a novel hierarchical skill transfer framework that integrates skill library with fine tactile representation, thereby combining high-level reasoning with low-level precise execution capabilities to efficiently transfer skills across long-horizon, contact-rich scenarios. Specifically, we first construct a skill library based on knowledge graph, using ``task graph'' and ``scene graph'' to represent task and scene information, respectively, and through ``state graph,'' complete the interaction between high-level task information and low-level precise physical information. On this basis, we build the hierarchical skill transfer framework, which divide the entire skill transfer process into three layers from high to low. At the task layer, we use LLMs, inputting the skill library in code form, then implement subtask sequence transfer through a four-stage prompting paradigm combining in-context learning and chain-of-thought prompting. At the motion layer, with the aid of precise spatial scene physical information from the skill library, we use the heuristic A* path planning algorithm \cite{xiangrong2021improved} to achieve adaptive trajectory transfer. Finally, at the physical layer, we propose an adaptive contour extraction and posture perception method based on tactile representation \cite{xu2024unit}, dynamically acquiring high precision contour and posture information through visual-tactile images, adjusting contact position and posture parameters to ensure the efficiency of skill transfer in contact-rich new environments.

Our contributions are summarized as follows:
\begin{itemize}
  \item A hierarchical skill transfer framework based on skill library and tactile representation is proposed, incorporating high-level reasoning and low-level precise execution to efficiently adapt existing skills to new scenarios.
  
  \item A knowledge graph-based skill library construction method is proposed, which represents task and scene information through ``task graph'' and ``scene graph'' and utilizes ``state graph'' to facilitate their interaction.
  
  \item A tactile perception-based adaptive contour extraction and pose estimation method is proposed, enabling internal self-adaptation of transferred skills at the physical level to ensure their effectiveness in new scenes.
\end{itemize}

\section{ RELATED WORK}

\subsection{Skill Library}
A robot skill library is a structured set of skills, each corresponding to a specific task’s manipulation process. These libraries are built through Imitation Learning and Reinforcement Learning, aimed at enabling skill reuse and composition for improved multi-tasking. Earlier robotic systems were task-specific with limited generalization. To improve adaptability, researchers proposed generalized skill libraries using modular skills for task planning \cite{kroemer2021review}. These libraries include basic and complex skills and can be categorized as task-specific or general \cite{rozo2013robot}. Skills are often represented using behavior trees or finite state machines \cite{colledanchise2018behavior}, and transfer learning allows robots to apply skills to new tasks.

\subsection{Skill Transfer}
Skill transfer reduces training time and sample requirements by transferring knowledge from a source to a target domain. Key approaches include transfer learning, multitask learning, and meta-learning. Sun et al. \cite{achituve2021self} proposed a self-supervised method for robots to adapt to dynamic environments. Yu et al. \cite{yu2018learning} demonstrated skill transfer through similar task examples, improving a robot's object grasping ability. Multitask learning boosts efficiency across tasks by sharing representations \cite{du2018direct}. Meta-learning, like MAML, helps robots adapt quickly with few samples \cite{finn2017model}. with transfer learning, such as Progressive Neural Networks (PNNs) by Rusu et al. \cite{rusu2016progressive}, shows strong skill transfer in complex tasks.

\subsection{Tactile Perception}
Li et al. \cite{li2014localization} achieved precise tracking and positioning of small widgets through high-precision matching and map analysis, successfully locating and inserting USB connectors. Izatt et al. \cite{izatt2017tracking} extended this approach by proposing a filtering algorithm that fuses vision and tactile information to track object poses accurately, enabling precise grasping and insertion operations for a screwdriver. Both studies assumed known object geometries and focused on model-based tracking. Daolin Ma et al. \cite{ma2021extrinsic} addressed the challenge of locating contact points with unknown objects and environments by directly solving constraint problems to estimate the contact positions. Yu She et al. \cite{she2021cable} used thresholding and PCA to extract the principal components of the contact area depth map, determining cable orientation for manipulation tasks. Achu Wilson et al. \cite{wilson2023cable} combined tactile-guided low-level motion control with vision-based task parsing to achieve cable routing and assembly tasks.

\begin{figure*}[!t]
	\centering
	\includegraphics[width=\textwidth]{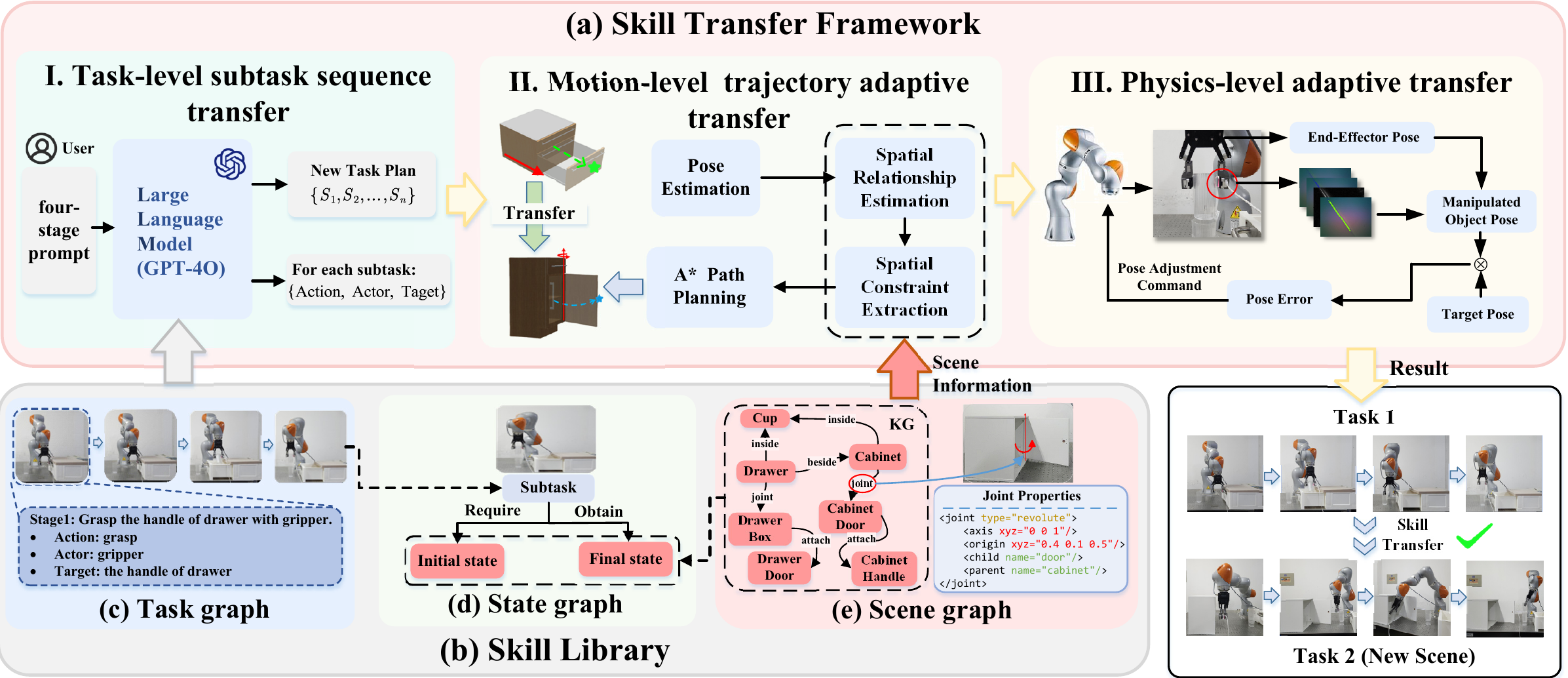} 
	\caption{\textbf{Overview Framework.} We first propose the skill library based on knowledge graph, representing high-level tasks and low-level scene information through ``task graph'' and ``scene graph''. The ``state graph'' facilitates interaction between them via ``require'' and ``obtain'' mechanisms. Based on this, we propose hierarchical skill transfer framework. At the task level, we employ a four-stage prompting paradigm based on large language models and the skill library to achieve subtask sequence transfer. At the motion level, we retrieve relevant physical information from the ``scene graph'' and utilize heuristic path planning algorithms to achieve adaptive trajectory transfer. At the physical level, we introduce a tactile perception-based adaptive posture adjustment method, dynamically modifying contact positions and postures through contour extraction and pose sensing.} 
	\label{structure} 
\end{figure*}

\section{METHOD}
Our proposed method consists of two parts: Skill Library constructed based on knowledge graph and Hierarchical Skill Transfer Framework, as shown in Figure~\ref{structure}. Here we will discuss in detail: (1) How to construct robotic manipulation skill library based on the knowledge graph (Sec. III-A)? (2) How to achieve hierarchical skill transfer based on the skill library and tactile representation (Sec. III-B)? (3) How is tactile representation implemented and how does it function in physical layer transfer (Sec. III-C)?

\subsection{The Construction of Skill Library}
To enable robots to understand high-level semantic skills and detailed scene information at the physical level, we constructed skill library based on knowledge graph(KG) \cite{daruna2019robocse}. Specifically, the skill library consists of three parts: the ``Task Graph,'' which hierarchically constructs knowledge by decomposing long-horizon tasks into sub-task sequences; the ``Scene Graph,'' which is built to effectively describe precise physical information and spatial constraints of manipulation objects; and the ``State Graph,'' which is constructed to enable the interaction between the former two. In this section, we detail how each of these three components is constructed and how they interact to form a complete skill library.

\textbf{The Construction of Manipulation Task Graph.}
As shown in Figure~\ref{structure} (c), given a long-horizon, contact-rich manipulation task $T$ (e.g., retrieving a cup from drawer and stacking), we first decompose the task into multiple stages, forming a sub-task sequence $S = \{S_1, S_2, ..., S_n\}$, where each stage $S_i$ can be formalized as $S_i = \{A_i, O_i^{\text{actor}}, O_i^{\text{target}}\}$. In this, $A_i$ represents the action to be performed (e.g., grasping, opening drawer), $O_i^{\text{actor}}$ denotes the object initiating the action $A_i$, and $O_i^{\text{target}}$ denotes the object that the action $A_i$ is aimed at. For example, for the ``stacking'' action, the cup grasped by the gripper is the actor object, while the cup placed on the table is the target object.

Based on this definition, we decompose long-horizon tasks into two hierarchical levels: the task level and the motion level. The task level comprises the high-level manipulation task $T$ and the planned sequence of subtasks $S = \{S_1, S_2, ..., S_n\}$. The action level consists of an action primitive sequence $\{A_1, A_2, ..., A_n\}$, where action primitives are low-level, short-term actions that can be directly executed by robots through motion planning \cite{xiangrong2021improved}. This hierarchical design distinguishes high-level semantic information from low-level executable actions, providing a foundation for hierarchical skill transfer in robotics.

Specifically, we utilize Neo4j to construct ``Task Graph'' $G^T = (V^T, E^T)$, where subtasks at the task level and action primitives at the motion level are represented as nodes $V^T$ in the graph, and the relationships between these nodes are represented as edges $E^T$. Constructing the knowledge graph with Neo4j allows each node $V^T$ and edge $E^T$ to have their own properties. We incorporate $O_i^{\text{actor}}$ and $O_i^{\text{target}}$ from $S_i$ as attributes of the corresponding node $V^T$. The properties of the edges $E^T$ represent different execution relationships within the task and action levels, such as ``start,'' ``next,'' and ``end''. Additionally, edges between the task and action levels encode ``contain'' relationship, indicating that a subtask $S$ may contain multiple sequentially executed actions $A_i$.

\textbf{The Construction of Manipulation Scene Graph.}
We represent the scene graph as \( G^S = (V^S, E^S) \), where \( V^S = \{v^S_1, v^S_2, \dots, v^S_n\} \) denotes the set of objects in the scene, and each object node \( v^S_i \) includes attributes such as category, position, and orientation. The edges \( E^S = \{e^S_1, e^S_2, \dots, e^S_m\} \) define the spatial or functional relationships between objects.

As shown in Figure~\ref{structure} (e), when constructing the scene graph to guide robot manipulation, we extend these concepts. Instead of modeling only entire objects as nodes $v_i^S$, represent the constituent parts of objects as nodes $v_i^S$. This is because robotic manipulation requires not only an understanding of an object's overall spatial properties but also its operable parts, such as the handle of a drawer or cabinet. Each node is assigned three attributes: category $c_i$, position $p_i$ (Cartesian coordinates), and orientation $q_i$ (quaternion). Consequently, the edges $e_i^S$ represent not only spatial relationships between objects (e.g., inside, beside) but also functional relationships between parts of an object (e.g., joint, attach).

When an edge $e_i^S$ represents spatial relationship, its attributes describe the relative position and orientation between two objects. If $e_i^S$ represents a joint connection, additional joint attributes are included, as an object's movement constraints are crucial for robotic manipulation. These joint attributes are encoded using XML, a common language for spatial modeling. For example, as depicted in Figure~\ref{structure} (e), the cabinet body and door are connected by a revolute joint, with the rotation axis defined as vertical (xyz=``0 0 1''), the joint location at the ``origin,'' and the connected parts labeled as ``child'' and ``parent.''

\textbf{The Construction of Manipulation State Graph.} 
Although the ``task graph'' and ``scene graph'' are constructed, they remain independent of each other. To establish a connection between them, we introduce the concept of ``state graph,'' which associates the subtask sequence in the ``task graph'' with the entity relationships in the ``scene graph'', as illustrated in Figure~\ref{structure} (d). To enable their interaction during robot execution, we define the concepts of ``Require'' and ``Obtain.'' The ``Require'' mechanism specifies the necessary conditions for executing a given subtask $S_i$, including the initiating object $O_i^{\text{actor}}$, the target object $O_i^{\text{target}}$, and the required state properties. These properties can be extracted from the attributes of nodes $V^S$ and edges $E^S$ in the skill library knowledge graph. The ``Obtain'' mechanism captures the state changes resulting from the completion of $S_i$. These changes are then fed back into the ``scene graph'', updating the attributes of relevant nodes and edges, thereby providing information support for the subsequent subtask execution. This entire process can be implemented using Neo4j.

\subsection{Hierarchical Transfer Framework of Manipulation Skills}
Our framework achieves hierarchical skill transfer based on skill library and tactile representation. Specifically, it sequentially implements task-level subtask sequence transfer, motion-level trajectory adaptive transfer, and physical-level manipulation adaptive transfer based on tactile representation. In summary, this framework enables skill transfer across different scenarios at the task level and achieves adaptive goal accomplishment strategies within skills at the motion and physical levels.

\textbf{Task-level Subtask Sequence Transfer.} In similar task scenarios, the subtask sequences required for planning and execution are often similar. For example, the robot skill sequence for opening a drawer and opening a door typically involves: approaching, grasping, pulling, and releasing. However, it is important to note that there are differences in the objects being manipulated and the modes of motion. To facilitate the transfer of subtask sequences between similar scenarios, the robot must fully understand both the transfer task and the transfer scenario. Therefore, we introduce the LLM, leveraging contextual learning and chain-of-thought prompting, to construct a new four-stage prompting paradigm. This approach utilizes the LLM's powerful reasoning and generalization capabilities to achieve this objective.

First, we provide the LLM with a comprehensive description of the complex planning task and the ``state graph'' descriptions from the skill library. This includes detailed accounts of long-horizon tasks, their subtask sequences, and the meanings and specific components associated with ``Require'' and ``Obtain'' in the ``state graph''.

Next, we enable the LLM to comprehend the entire knowledge graph of the operations skill library. We input the skill library into the LLM as Neo4j-based code, which helps the model understand subtask sequences in the task graph, components of objects in the scene graph, their connections, and their relative positional relationships.

In the third step, we guide the LLM to deeply understand the spatial semantic information within the scene graph. Due to the complexity of the scene graph, although the LLM can effectively analyze the attributes of various components within the scene graph after the first two steps, its understanding of spatial semantic information remains insufficient. For instance, it struggles to accurately output the spatial relationships of the components within objects. To address this, we build upon the second step by designing a specific triple template ``Object1-Connection Type-Object2'' to enhance the prompts. Here, ``Object'' represents objects or their components, while ``Connection Type'' corresponds to the spatial or functional relationship between the two objects.

Finally, reference plan for known complex tasks and a description of similar tasks to be executed in the new scenario are provided. In this step, we impose constraints through ``note'' to ensure that the subtask sequence transfer remains within the defined boundaries, thus preventing the generation of redundant information. As a result, the LLMs outputs a task planning sequence $S = \{S_1, S_2, ..., S_n\}$ for the new scenario, where each $S_i$ corresponds to an action $A_i$. For each action $A_i$, the initiating object in the new scenario is $O_i^{\text{actor}}$, and the target object is $O_i^{\text{target}}$, represented as $\{A_i, O_i^{\text{actor}}, O_i^{\text{target}}\}$.

\textbf{Motion-level Adaptive Trajectory Transfer.} After task-level transfer, the subtask sequence is represented using human semantic knowledge, which the robot cannot directly execute. Therefore, motion-level trajectory transfer is necessary. Different task scenarios involve distinct object motion patterns; for example, pulling a drawer requires linear motion along an axis, whereas opening a cabinet involves rotational motion around an axis. Additionally, obstacles may differ across scenarios, potentially leading to collisions. To address this, we propose an adaptive trajectory transfer method that utilizes heuristic A* algorithm-based trajectory planning in conjunction with skill library.

Once interaction primitives and their corresponding spatial constraints are defined for each subtask, the subtask execution can be framed as an optimization problem. For each action $A_i$ in the subtask sequence $S_i$, we first retrieve state information, including positions, orientations, and joint attributes of $O_i^{\text{actor}}$ and $O_i^{\text{target}}$, from the scene graph using the ``Require'' query based on task-level transfer results $\{A_i, O_i^{\text{actor}}, O_i^{\text{target}}\}$. Subsequently, the start and end points of action $A_i$'s trajectory are computed, and a corresponding 3D discretized map is generated. The A* algorithm is then applied for path planning. To prevent collisions between the end-effector and obstacles in the environment when operating in a new scenario, we define the collision loss $L_{\text{collision}}$ as:
\begin{equation}
L_{\text{collision}} = \sum_{i=1}^{N} \max(0, \delta - d(p_e^i, O_i))^2
\end{equation}
where $d(p_e^i, O_i)$ denotes the distance between the end-effector $p_e^i$ and the obstacle $O_i$, while $\delta$ represents the minimum allowable safety distance. When applying the A* algorithm for path planning, minimizing collision loss ensures the successful execution of the action sequence $A_i$ corresponding to the subtask sequence $S_i$ in a new task environment. This strategy not only prevents collisions but also promotes smooth motion, facilitating the adaptive transfer of motion-level trajectories to the new environment.

\begin{figure}[!t]
	\centering
	\includegraphics[width=6.5cm]{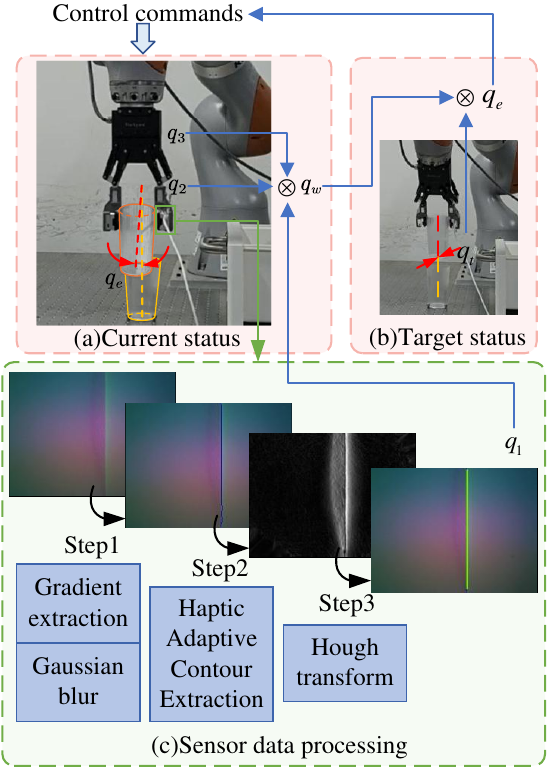} 
	\caption{In Figure (a), \( q_2 \) and \( q_3 \) correspond to the sensor's mounting pose and the end-effector pose of the robotic arm, respectively, while Figures (a) and (b) illustrate the pose error \( q_e \) that needs to be adjusted and the target pose \( q_t \) to be reached, represented by the yellow and red dashed lines, respectively. In Figure (c), the pose information \( q_1 \) obtained through the algorithm is shown, with the blue scatter points indicating the extracted edges and the green lines representing the directions derived from the Hough transform. Ultimately, by performing calculations with the target pose, a pose control command for the robotic arm is generated.} 
	\label{tactile_perception} 
\end{figure}

\textbf{Physical-level Transfer based on Tactile Representation.}
Task-level and motion-level transfers facilitate the generalization of specific tasks in new scenes. However, for tasks involving intricate tactile interactions, robots must manage the variability of object attributes and the complexity of underlying contact mechanisms, which constrains the adaptability and generalization of their skills.

We introduce an adaptive learning mechanism that dynamically adjusts manipulation strategies based on real-time feedback from high-resolution visuotactile sensors such as Gelsight. The feedback encompasses information about an object's contours, textures, posture, forces, and torques. For example, when grasping objects with varying textures, robots can modify their force and grip angle in response to perceived hardness or slip characteristics, thereby maintaining stability and precision under different contact conditions and adapting to diverse operational environments.

However, existing tactile perception methods struggle to maintain robustness across diverse scenarios. To achieve physical-level tactile perception-driven adaptive transfer, we propose an adaptive method for contour extraction and pose estimation based on tactile perception, which will be introduced in the following section (Sec. III-C).

\subsection{Pose Adjustment Utilizing Tactile Contour Extraction}
In order to predict the pose of textured objects, we propose dividing the tactile image into three regions: texture region, contour region, and non-contact region. Pose information is then extracted from the texture region to facilitate accurate pose prediction. In a typical tactile image, the contour region is characterized by sharp gradient changes; the non-contact region primarily exhibits scattered, low-gradient noise; and the texture region is determined by the surface and contact conditions of the touched object, typically displaying regular patterns with moderate gradient strength.

The algorithm first defines the texture threshold ($T_{texture}$) as the floor value of the average of non-zero gradient values in the gradient map $G$, given by the following equation:
\begin{equation}
T_{texture} = \left| \frac{1}{|G'|} \sum_{i,j:G_{i,j} \neq 0} G_{i,j} \right| \label{eq:texture}
\end{equation}

Next, each pixel $(i,j)$ in the image $I$ is smoothed by applying a Gaussian kernel $G_k$, resulting in the smoothed image $S$. Then, the gradient magnitude $G_{i,j}$ and gradient direction $\Theta_{i,j}$ are computed for each pixel. Non-maximum suppression is applied to retain local maximum gradient values, refining the contour points. A dual-threshold detection is then performed on these contour points using dynamically adjusted high and low thresholds ($T_{\text{high}}$ and $T_{\text{low}}$), optimized to match the texture information. Through this process, the final set of contour points $C$ is determined. Finally, the Hough Transform is used to extract lines from the contour point set and compute the line direction vectors $H$, providing the necessary data for further processing or analysis.

\begin{figure*}[!t]
	\centering
	\includegraphics[width=15.8cm]{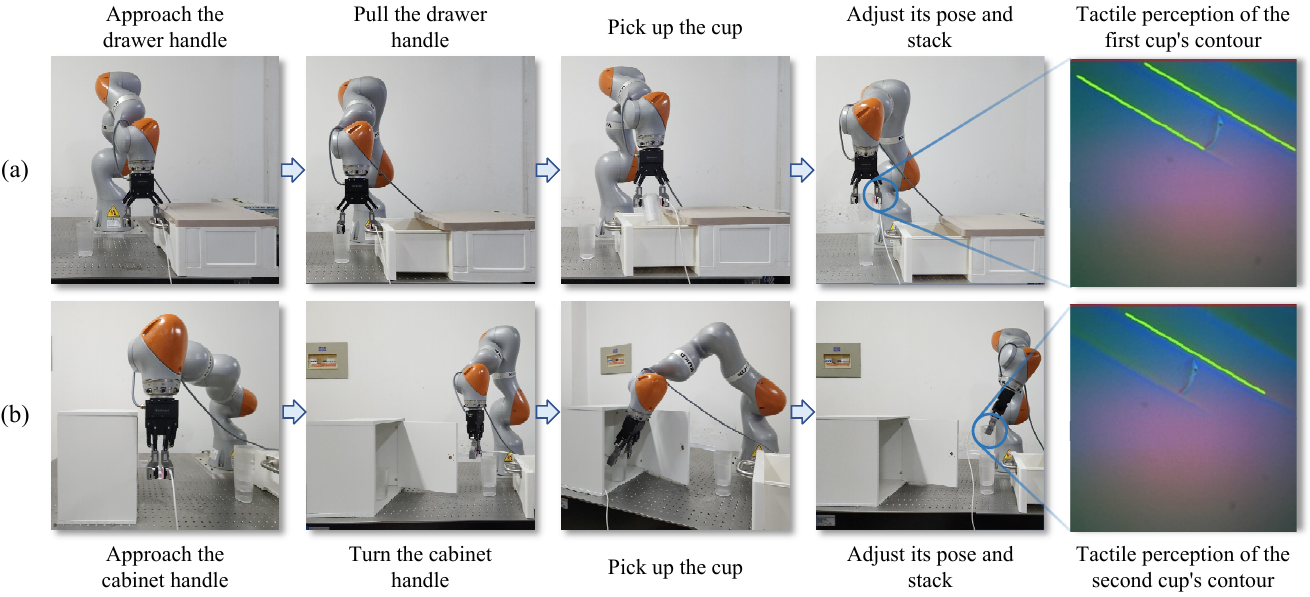} 
	\caption{\textbf{Skill Transfer Result.}  Figure (a) illustrates the initial skills in the skill library, while Figure (b) depicts the process of successfully completing the task in new scene after skill transfer. During this process, tactile feedback is integrated into the control loop.} 
	\label{skill_transfer_experiment} 
\end{figure*}

\begin{algorithm}
	\caption{Tactile Contour Extraction Algorithm}
	\label{alg:AOS}
	\begin{algorithmic}[1] 
		\State \textbf{Input:} A 320x240 Gelsight image $I$
		\State \textbf{Output:} Contour point set $C$ and line direction vectors obtained by Hough transform $H$
		\Procedure{ContourExtraction}{}
		\State $T_\textit{texture}$ $\gets$ \{$G$\} with equation \ref{eq:texture}
		\For{each pixel $(i, j)$ in $I$}
		\State $S_{i,j} \gets$ smooth $I$ with $G_k$
		\State $G_{i,j} = \sqrt{(S_{i+1,j} - S_{i-1,j})^2 + (S_{i,j+1} - S_{i,j-1})^2}$
		\State $\Theta_{i,j} = \arctan\left(\frac{S_{i,j+1} - S_{i,j-1}}{S_{i+1,j} - S_{i-1,j}}\right)$
		\EndFor
		\For{each pixel $(i, j)$ in $G$}
		\State $N_{i,j} \gets$ \{$G$, $\Theta$\} non-maximum suppression 
		\State $C_{i,j} \gets$ \{$G$, $T_\textit{texture}$\} dual threshold detection 
		\EndFor
		\State $T_{\text{high}}, T_{\text{low}} \gets$ adjust with $T_\textit{texture}$
		\State $H \gets$ \{C\} with Hough Transform
		\EndProcedure
		\State \textbf{Return:} $C$ and $H$
	\end{algorithmic}
	\label{algorithm}
\end{algorithm}

The tactile contour extraction algorithm based on image texture is designed to process a 320x240 resolution Gelsight image and extract contour point sets and line direction vectors using the Hough transform. The complete processing flow and pseudocode are presented in Figure~\ref{tactile_perception} and Algorithm~\ref{algorithm}, respectively.

After obtaining the object's pose relative to the tactile sensor \( q_1 \), its pose in the world coordinate system \( q_w \) is computed as:
\begin{equation}
	q_w = q_3 \otimes q_2 \otimes q_1
\end{equation}
where \( q_2 \) and \( q_3 \) represent the sensor's installation pose relative to the end-effector and the end-effector's pose in the world frame, respectively.
Then the error quaternion \(q_e\) is computed as
\begin{equation}
	q_e = q_t \otimes q_w^{-1}
\end{equation}
representing the adjustment required for the robotic arm to align the grasped object with the desired pose \(q_t\), thereby providing the command for end-effector correction and improving grasping accuracy and stability.

\section{EXPERIMENTS}
\subsection{Experimental Setup}
\textbf{Hardware Configuration.} Our experimental platform is built around a KUKA LBR iiwa14 robotic arm. The tactile sensor employed is the Gelsight mini, with sampling frequency of 10Hz. For low-level control, we use position control in the end-effector's Cartesian space at frequency of 100 Hz, enabling the generation of smooth interpolated trajectories.

\textbf{Implement Details.} We use Neo4j to build the knowledge graph-based skill library. We use GPT-4O from the OpenAI API as the Large Language Model. Minimal few-shot prompts are employed to help the LLM understand its role. Additionally, the chain-of-thought technique is used to facilitate deeply understanding of the scene by the LLM. Specific prompts used can be found in the code repository.

\subsection{Validation of the Effectiveness of the Hierarchical Skill Transfer Framework}
In our experiment, we set up a drawer and a cabinet with a door, both containing vertically striped cups placed in different positions and orientations. The robot, equipped with a skill library, has already learned to open the drawer, retrieve the cup, move it to a designated location, and complete a stacking task. However, the robot is now presented with a similar but previously unseen scenario, where the skill library does not provide a complete task-specific guidance. Our goal is to enable skill transfer, allowing the robot to autonomously infer the necessary actions in the new environment. Specifically, the robot must open the cabinet door, which involves rotational rather than linear movement, retrieve cups from various positions and orientations, and transport them to the designated location for stacking. In this transfer scenario, the robot is required to complete a sequence of complex sub-tasks, including approaching, grasping, pulling, and releasing the handles of drawers or cabinet doors, followed by approaching, picking up, and moving to a designated position for precise cup stacking. Throughout the process, the robot must adaptively perform trajectory planning tasks based on scene information provided by the skill library. During the cup stacking phase, the robot will adjust its posture adaptively, utilizing tactile perception of physical-level information, such as the contour and orientation of the cup, to successfully complete the stacking task.

As shown in Figure~\ref{skill_transfer_experiment}, experimental results demonstrate that the robot successfully completed the task in the novel environment without prior training, validating the effectiveness of the proposed hierarchical skill transfer framework in handling long-horizon, contact-rich manipulation tasks.

\subsection{Evaluation of the Skill Transfer Framework}

\textbf{Baselines.} We compare our method with direct policy transfer method based on reinforcement learning. In complex scenarios, skill transfer and generalization have been key focuses of previous research, primarily achieved through imitation learning from demonstrations and reinforcement learning in simulation \cite{xia2024kinematic}. Imitation learning, however, requires a large amount of data for model pre-training.Therefore, here we choose to compare with a direct policy transfer method based on reinforcement learning. The simulation environment is constructed using Robosuite, an open-source framework for robot simulation and control.

Specifically, the direct policy transfer method refers to using the SAC reinforcement learning algorithm \cite{haarnoja2018soft} for training in a simulated scenario in advance. The learned policy network from the SAC algorithm is then directly applied to an untrained door-opening task to guide the robot arm's movements. The design of the reinforcement learning reward function is as follows:
\begin{equation}
r = \begin{cases} 
      2 & \text{success} \\
      0.25 \times (1 - \tanh(10 \times d)) & \text{unsuccess} 
   \end{cases} 
\end{equation}

where \textit{success} and \textit{unsuccess} indicate whether the task of opening the drawer is completed, and $d$ represents the Euclidean distance between the robot gripper and the target position on the drawer handle.

\begin{figure}[!ht]
	\centering
	\includegraphics[width=6.5cm]{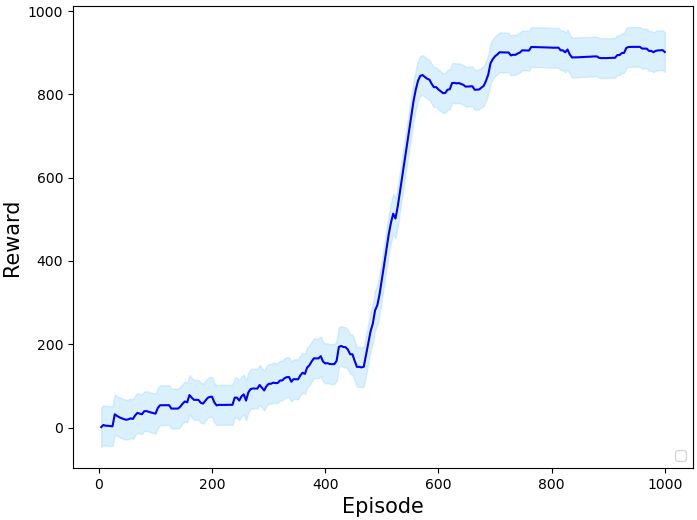} 
	\caption{Results of training the SAC algorithm on the drawer-opening task in simulation environment. } 
	\label{open_drawer} 
\end{figure}

Training is conducted using the SAC algorithm from stable baselines3 for 1000 episodes, with each episode consisting of a fixed 1000 time steps. The training results are shown in the Figure~\ref{open_drawer}, indicating that the reinforcement algorithm has learned the current skill. The trained policy network from the SAC algorithm is directly transferred and applied to the door-opening task. Ten episodes are tested, each with 1000 steps, while our method is also tested for 10 episodes in the same environment.

\begin{table}[htbp]
\centering
\renewcommand{\arraystretch}{1.5} 
\caption{Skill Transfer Method Success Rate Comparison}
\label{tab:success_rate_comparison}
\begin{tabular}{lcc}
\hline
Method & Direct policy transfer & \textbf{Ours} \\
\hline
Suc. Rate & 0/10 & \textbf{8/10} \\
\hline
\end{tabular}
\end{table}

As shown in the TABLE~\ref{tab:success_rate_comparison}, compared to the direct policy which struggles to successfully complete tasks in new scenarios, our skill transfer method can maintain a high success rate in new environments. This is attributed to our skill library and hierarchical skill transfer framework, which can understand tasks and scenarios, and adaptively adjust the execution strategy according to the new scenes during the transfer process. In contrast, policy transfer based on reinforcement learning merely involves transferring actions, making it difficult to adapt to complex and dynamic new scenarios.

\subsection{Verification of Adaptive Tactile Threshold Algorithm}
The goal of this experiment is to evaluate the effectiveness of the object pose perception module in various scenarios, with a focus on the accuracy and reliability of extracting six typical contours. To achieve this, we employ our adaptive method and compare its performance with a fixed-threshold edge extraction method.

\begin{figure}[htbp]
	\centering
	\includegraphics[width=\linewidth]{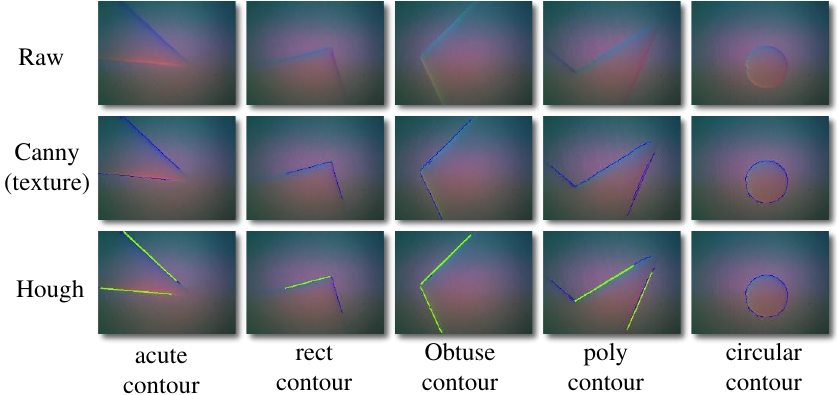} 
	\caption{Tactile adaptive contour extraction results. The blue scatter points represent the extracted contour points, while the green lines represent the straight edges perceived through the Hough line transform applied to the contour points.}
	\label{fig:experiment_3}
\end{figure}

During the experiment, multiple data collection sessions were conducted for each sample at a sampling rate of 10 Hz, with each session lasting approximately 5 seconds. To assess the differences in performance between the adaptive contour extraction algorithm and the fixed-threshold method, we utilized Root Mean Square Error (RMSE) to measure the positional error between the extracted contours and the ground truth. A lower RMSE indicates better algorithm performance. Figure~\ref{fig:experiment_3} shows the results of our algorithm intuitively and Table~\ref{tab:rmse-errors} summarizes the RMSE positional errors under different conditions for both algorithms.

\begin{table}[htbp]
	\centering
	\caption{RMSE of Tactile Adaptive and Fixed Thresholds on Various Geometrical Shapes.}
	\fontsize{9pt}{11pt}\selectfont
	\begin{tabular}{l>{\centering\arraybackslash}p{1.1cm} >{\centering\arraybackslash}p{1.1cm} >{\centering\arraybackslash}p{1.1cm} >{\centering\arraybackslash}p{1.1cm}} 
		\toprule
		& Rect & Acute & Right & Circle \\
		\midrule
		Fixed Threshold & 39.31 & 101.76 & 112.81 & 326.04 \\
		\textbf{Tactile Adaptive} & \textbf{37.15} & \textbf{61.52} & \textbf{62.38} & \textbf{63.80} \\
		\bottomrule
	\end{tabular}
	
	\vspace{0.2em}
	
	\begin{tabular}{l>{\centering\arraybackslash}p{1.1cm} >{\centering\arraybackslash}p{1.1cm} >{\centering\arraybackslash}p{1.1cm} >{\centering\arraybackslash}p{1.1cm}} 
		\toprule
		& \multicolumn{2}{c}{Pentagon} & \multicolumn{2}{c}{Line Bundle} \\
		\cmidrule(lr){2-3} \cmidrule(lr){4-5}
		& General & Edge & General & Edge \\
		\midrule
		Fixed Threshold & 123.30 & 203.59 & 117.44 & 214.16 \\
		\textbf{Tactile Adaptive} & \textbf{61.87} & \textbf{102.05} & \textbf{103.03} & \textbf{193.92} \\
		\bottomrule
	\end{tabular}
	\\[5pt]
	\parbox[t]{0.5\textwidth}{\raggedright Rect, acute, right represent straight line, acute angle, right angle respectively. A smaller numerical value indicates better performance.}
	
	\label{tab:rmse-errors}
\end{table}

The experimental data demonstrate that the adaptive contour extraction algorithm outperforms the fixed-threshold method when handling various geometric shapes and complex surfaces. Specially, the adaptive algorithm exhibits stronger robustness in scenarios with multiple edges (complex geometries) or contours located at the edge of the sensor’s field of view (challenging contact conditions). Figure~\ref{fig:experiment_3} and the error data clearly demonstrate that our algorithm can effectively perceive the object's pose information, providing valuable insights for further manipulations.

\section{CONCLUSIONS AND LIMITATIONS}
In this work, we propose hierarchical skill transfer framework based on skill library and tactile representation, which efficiently enables skill transfer and generalization across various new scenarios. We construct a skill library based on a knowledge graph, using ``task graphs'' and ``scene graphs'' to represent task and scene knowledge, and ``state graphs'' for interaction. Building on this foundation, we incorporate the skill library to achieve high-level reasoning and subtask transfer at the task level with the help of LLM, adaptive trajectory transfer at the motion level using heuristic path planning algorithms, and adaptive posture adjustment at the physical level based on tactile representation. Real-world experiments have also demonstrated the generalization capability of our framework in practical scenarios.

\textbf{Limitations.} While advantageous, our framework also has limitations. Due to the posture representation and tactile contour extraction algorithm, it cannot model deformable objects. Secondly, the currently used LLM is pre-trained on large-scale 2D images and lacks a true understanding of the 3D physical world, which sometimes leads to inaccuracies in motion path generation. This limitation hinders its ability to perform accurate spatial reasoning. Incorporating 3D inputs, such as point clouds, into the fine-tuning process of LLM may alleviate this challenge.







\bibliographystyle{BibTex/IEEEtran}
\bibliography{References}\

\end{document}